\documentclass[10pt,journal]{IEEEtran}

\usepackage{graphicx}
\usepackage{epstopdf}
\usepackage{subcaption}
\usepackage{algorithm}
\usepackage{algorithmic}
\usepackage{booktabs}
\usepackage{amsfonts,amssymb}
\usepackage{amsmath}
\usepackage{tcolorbox}
\tcbuselibrary{breakable}
\usepackage[graphicx]{realboxes} 
\usepackage[colorlinks=true, linkcolor=black, urlcolor=black, citecolor=black]{hyperref}
\renewcommand{\textcolor}[2]{#2} 

\ifCLASSINFOpdf
\else
\fi
\begin{document}
%

\title{Velocity and Density-Aware RRI Analysis and Optimization for AoI Minimization in IoV SPS}

\author{Maoxin~Ji,~Tong~Wang,~Qiong~Wu,~\IEEEmembership{Senior~Member,~IEEE},~Pingyi~Fan,~\IEEEmembership{Senior~Member,~IEEE},\\~Nan~Cheng,~\IEEEmembership{Senior Member,~IEEE},~and~Wen~Chen,~\IEEEmembership{Senior~Member,~IEEE}
\thanks{Maoxin Ji, Tong Wang and Qiong Wu are with the School of Internet of Things Engineering, Jiangnan University, Wuxi 214122, China (e-mail: maoxinji@stu.jiangnan.edu.cn, tongwang@stu.jiangnan.edu.cn and qiongwu@jiangnan.edu.cn).}
\thanks{Pingyi Fan is with the Department of Electronic Engineering, State Key laboratory of Space Network and Communications, Beijing National Research Center for Information Science and Technology, Tsinghua University, Beijing 100084, China (e-mail: fpy@tsinghua.edu.cn).}
\thanks{Nan Cheng is with the State Key Lab. of ISN and School of Telecommunications Engineering, Xidian University, Xi'an 710071, China (e-mail: dr.nan.cheng@ieee.org).}
\thanks{Wen Chen is with the Department of Electronic Engineering, Shanghai Jiao Tong University, Shanghai 200240, China (e-mail: wenchen@sjtu.edu.cn).}
}

\markboth{IEEE Communications Letters,~Vol.~xx, No.~xx, May~2025}%
{Shell \MakeLowercase{\textit{et al.}}: Bare Demo of IEEEtran.cls for Computer Society Journals}

\maketitle

\begin{abstract}
Addressing the problem of \textcolor{red}{Age of Information} (AoI) deterioration caused by packet collisions and vehicle speed-related channel uncertainties in Semi-Persistent Scheduling (SPS) for the Internet of Vehicles (IoV), this letter proposes an optimization approach based on Large Language Models (LLM) and Deep Deterministic Policy Gradient (DDPG). First, an AoI calculation model influenced by vehicle speed, vehicle density, and Resource Reservation Interval (RRI) is established, followed by the design of a dual-path optimization scheme. The DDPG is \textcolor{red}{guided} by the state space and reward function, while the LLM leverages contextual learning to generate optimal parameter configurations. Experimental results demonstrate that LLM can significantly reduce AoI after accumulating a small number of exemplars without requiring model training, whereas the DDPG method achieves more stable performance after training.
\end{abstract}


\IEEEpeerreviewmaketitle

\vspace{-0.2cm}
\section{Introduction}

%
\IEEEPARstart{T}{he} Internet of Vehicles (IoV) is pivotal in enabling intelligent transportation systems \cite{[6], 6883711, [9]}. \textcolor{red}{Within IoV, Vehicle-to-Vehicle (V2V) communication enables direct end-to-end data exchange between vehicles, facilitating the transmission of critical information such as Basic Safety Messages (BSMs), which is a key technology to support intelligent driving \cite{7036784, [10], [111]}. Due to the dynamic nature of traffic environments, vehicular tasks require low latency and high reliability. The Age of Information (AoI) directly measures the time elapsed from the generation of a data packet to its reception, effectively reflecting the freshness of information. As such, AoI serves as a crucial performance metric in vehicular networks \cite{[7]}.}

\textcolor{red}{In the 5G New Radio (NR) Vehicle-to-Everything (V2X) standard, V2V communication typically adopts Mode 2 for autonomous resource selection, where Semi-Persistent Scheduling (SPS) is used to contend for and occupy resources \cite{[18], 9472822}. However, traditional SPS relies on empirically fixed parameters such as the Resource Reservation Interval (RRI), which may lead to suboptimal performance, especially under varying vehicle densities \cite{11054161}. \cite{11054234} analyzes how continuous resource occupancy in SPS impacts system AoI, showing that collisions causing consecutive transmission failures degrade AoI performance. To address this, \cite{[11]} theoretically and via Monte Carlo simulations study the expected peak AoI for different RRIs under varying levels of vehicle density. Moreover, \cite{9472822, 10355839} propose adaptive schemes to dynamically adjust the Resource Reselection Counter (RC) and RRI based on channel availability, reducing collisions and improving road safety. These works underscore the importance of adaptive parameter tuning in enhancing SPS performance.}

\textcolor{red}{Existing studies mainly focus on vehicle density when selecting the RRI. Under high density, limited channel resources increase the likelihood of collisions and consecutive transmission failures, leading to higher system AoI. In reality, high-speed vehicles—particularly those traveling in opposite directions—experience significant Doppler shifts, which may cause additional transmission failures. Furthermore, in the absence of configured Physical Sidelink Feedback Channel (PSFCH), broadcast messages such as BSMs typically use blind retransmission, where retransmission intervals are often tied to the RRI and directly affect the packet’s AoI \cite{10758027}. Therefore, RRI selection should consider both vehicle speed and density to optimize communication performance.}


\textcolor{red}{In this letter, we analyze the impact of vehicle speed, density, and RRI on the system AoI in highway scenarios by incorporating uncertainties in physical channel conditions induced by vehicle speed and the probabilities of resource packet collisions due to vehicle density and RRI}\footnote{\href{https://github.com/qiongwu86/AI-Based-IoV-Resource-Scheduling-SPS-Parameter-Optimization-for-AoI-Minimization}{The source code has been released at: https://github.com/qiongwu86/AI-Based-IoV-Resource-Scheduling-SPS-Parameter-Optimization-for-AoI-Minimization}}. To find the optimal parameter configuration, we leverage the predictive and contextual learning capabilities of Large Language Models (LLMs) to iteratively minimize the system AoI. We also address the problem using traditional Deep Reinforcement Learning (DRL) methods for comparison. Experimental results demonstrate that optimizing vehicle density, RRI, and vehicle speed significantly reduces AoI in IoV systems, thereby enhancing information timeliness.

\section{System Model}
\begin{figure}[h]
	\centering 
	\includegraphics[width=\columnwidth]{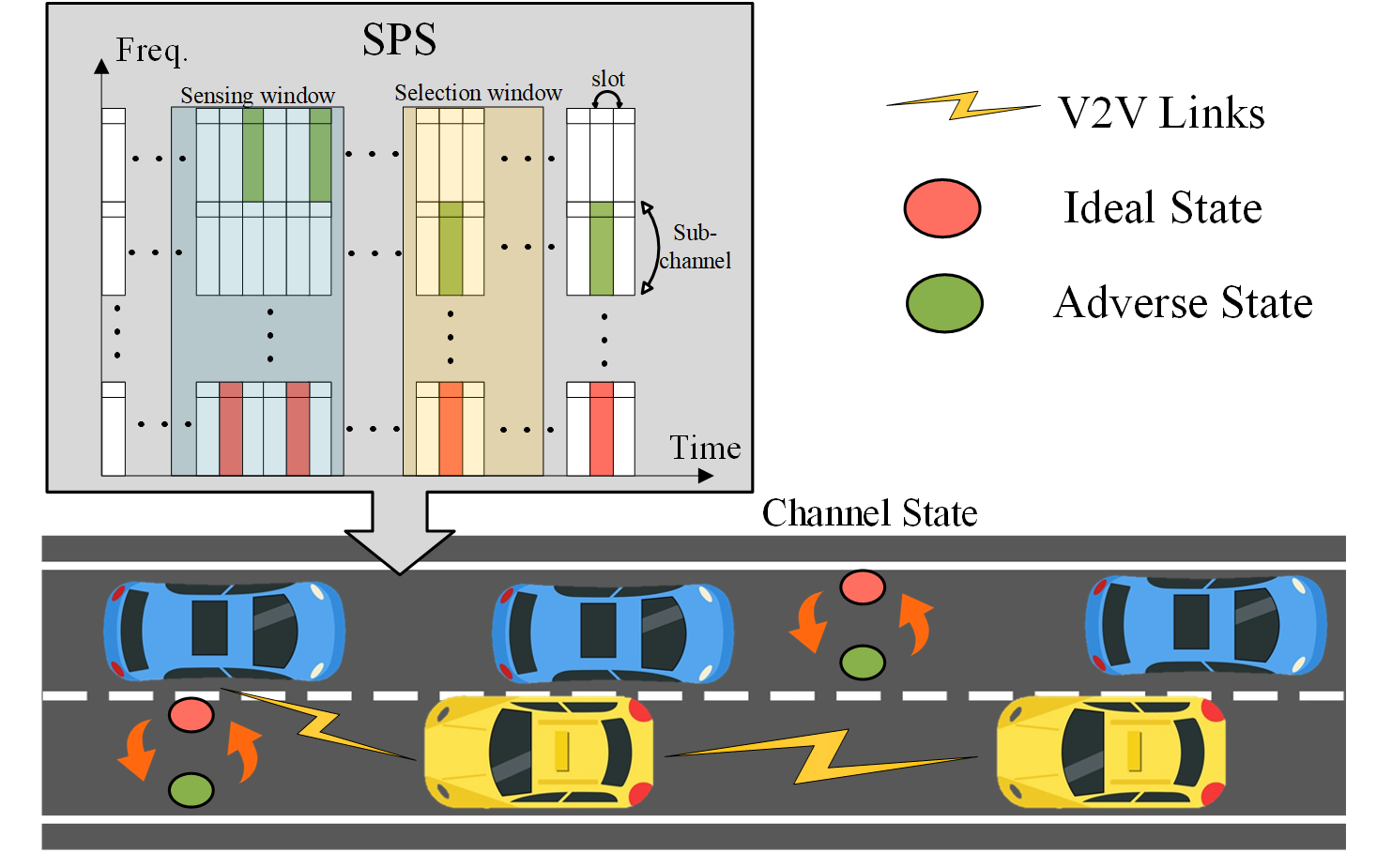}
	\caption{System model.}
	\label{fig:1}       
	\vspace{-0.5cm}
\end{figure}
\textcolor{red}{In this section, we consider a finite highway segment, as shown in Fig.~\ref{fig:1}, with two lanes carrying traffic in opposite directions. The highway spans the interval \([-L, L]\), where \(L\) is half its length. Let \(M = \{1, 2, \ldots, m_{\text{total}}\}\) be the set of vehicles communicating via V2V. Vehicles are uniformly distributed with density \(\rho_l\), so the total number of vehicles is:}

\begin{equation}\label{}
	m_{total} = 2 \rho_l L.
\end{equation}

\textcolor{red}{Although vehicle speed can be modeled as a random variable, traffic flow theory typically relates speed and density when describing highway traffic \cite{[14]}. In particular, Greenshields proposed a classical linear model that captures this relationship \cite{88995}.} Assuming a traffic flow \(Q\) on the highway, the speed can be expressed as:
\begin{equation}\label{}
	v = \frac{Q}{\rho_l}.
\end{equation}

\textcolor{red}{V2V communication based on NR V2X Mode 2 employs SPS to occupy and contend for resource blocks. Considering potential packet collisions and physical-layer transmission failures, when the PSFCH is not configured, vehicles perform blind retransmissions during the RRI with intervals of at least \( t_{\text{GAP}} \).}

Assuming each vehicle has a sensing range of \(R_s\), and assuming constant inter-vehicle spacing around vehicle \(n\), the number of vehicles within its sensing range of vehicle \(n\) can be calculated as:
\begin{equation}\label{}   
	N_s  = 2\rho_l R_s.
\end{equation}
According to \cite{[11]}, when considering the impact of hidden terminals, the number of hidden terminals for the \(m\)-th nearest transmitting vehicle to vehicle \(n\) is approximately \(m\). Therefore, the packet collision probability for vehicle \(n\) can be calculated as:
\begin{equation}\label{}   
	P_{\text{coll}} = 1 - \left(1 - \frac{1}{{N_r} - {N_s}/2}\right)^m,
\end{equation}
where,
\begin{equation}\label{}   
	N_r = {\frac{\mathrm{RRI}\cdot n_s}{t_s}},
\end{equation}
\textcolor{red}{where, \(N_r\) denotes the maximum number of Resource Block Groups (RBGs) that can be selected within the selection window for each vehicle, with \(1 \leq m \leq \frac{N_s}{2}\). \(n_s\) is the number of RBGs per time slot, and \(t_s\) is the duration of each time slot.}

\textcolor{red}{If, due to collisions, the target vehicle fails to successfully receive the data packet sent from the \(m\)-th nearest neighbor vehicle, blind retransmissions are also likely to encounter collisions, resulting in the retransmission being deferred to the next RRI period.} Assuming the hidden terminal effect occurs independently in each RRI, the additional delay of the packet from the \(m\)-th nearest neighbor vehicle caused by collisions follows a geometric distribution with parameter \(1 - P_{\text{coll}}\). The expected additional delay is given by:
\begin{equation}\label{}   
    E[T_a] = \sum_{i=1}^\infty i \cdot \mathrm{RRI} \cdot (P_{\text{coll}})^i (1 - P_{\text{coll}}) = \frac{\mathrm{RRI} \cdot P_{\text{coll}}}{1 - P_{\text{coll}}},
\end{equation}
where \(i\) denotes the \(i\)-th transmission attempt. Then the average queuing time required for the ${m^{th}}$ nearest vehicle to send a packet can be written as:
\begin{equation}\label{}   
	E[{T_m}] = RRI + \frac{{RRI \cdot P_{\text{coll}}}}{{1 - P_{\text{coll}}}}. 
\end{equation}
Since the queuing time depends on the relative position of the vehicle, the average queuing time over all vehicles can be written as:
\begin{equation}\label{eq1}   
	{T_q} = \frac{2}{{{N_s}}}\sum\limits_{m = 1}^{{N_s}/2} {E[{T_m}]}. 
\end{equation}

\textcolor{red}{After vehicular data transmission, the inherent randomness of wireless channels may cause transmission failures, which necessitate retransmission and thus cause data to re-enter the queue. Therefore, the packet loss probability directly impacts the queuing delay.}

Following the approach in \cite{8493300}, the channel state is modeled as a Markov process \textcolor{red}{where, depending on the signal-to-noise ratio (SNR), the channel transitions between an ideal state and an adverse state. Specifically, when the SNR exceeds \(\frac{E[\text{SNR}]}{F}\), where \(F\) denotes the fading margin, the channel is considered in an ideal state; otherwise, it is in an adverse state.}

At a vehicle speed \(v\) and carrier frequency \(f_c\), the Doppler frequency is given by \(\displaystyle f_d = \frac{f_c v}{c}\), where \(c\) is the speed of light. \textcolor{red}{The probability that a packet sent under adverse channel conditions is lost is given by \(p_e = 1 - e^{- \frac{1}{F}}\).} The correlation coefficient between channels in the adverse state at Doppler frequency \(f_d\) is \(\rho = J_0\left(\frac{2 \pi f_d}{\theta}\right)\), where \(J_0\) is the zero-order Bessel function \textcolor{red}{of the first kind}, and \(\frac{1}{\theta}\) \textcolor{red}{corresponds to} the packet transmission duration \textcolor{red}{over the channel}. This leads to the parameter \(\eta = \sqrt{\frac{2}{F(1 - \rho^2)}}\).

The \textcolor{red}{state transition probabilities} of the channel in the IoV system can thus be defined \textcolor{red}{according to a Markov model} as:
\begin{equation}
	\begin{aligned}
		&p_p = \frac{\mathbb{Q}(\rho \eta , \eta) - \mathbb{Q}(\eta , \rho \eta)}{e^{\frac{1}{F}} - 1} + 1, \\
		&p_i = \frac{1 - p_e (2 - p_p)}{1 - p_e},
	\end{aligned}
\end{equation}
where \(p_p\) denotes the probability that the channel remains in an adverse state, \(p_i\) denotes the probability it remains in an ideal state, \textcolor{red}{and \(\mathbb{Q}(\cdot, \cdot)\) represents the generalized Marcum Q-function.}

Assuming that a vehicle requires \(L\) link-layer frames to transmit a data packet, let \(l_L\) denote the probability that at least one frame transmission fails when the initial channel state is ideal, and \(v_L\) the analogous probability when the initial state is adverse. Since the first frame transmission always fails in the adverse state, we have \(v_L = 1\).

The recursive update for \(l_i\) is:
\begin{equation}
	l_i = p_i l_{i-1} + (1 - p_i) v_L,
\end{equation}
where \(l_i\) denotes the probability that at least one frame has failed up to the \(i\)-th frame.

Therefore, the overall packet loss probability during transmission can be expressed as:
\begin{equation}
	p_d = v_L \cdot \frac{1 - p_i}{2 - p_p - p_i} + \frac{l_L}{1 + \frac{1 - p_i}{1 - p_p}}.
\end{equation}

\textcolor{red}{Since a single data packet can only be transmitted after successfully completing SPS resource allocation, if the packet is lost during transmission, it must re-enter the SPS stage for retransmission. However, considering the blind retransmission mechanism in the 5G NR V2X standard, it is assumed that the sender will perform a blind retransmission at $RRI/2$ (no less than \(t_{\text{GAP}}\)). If the blind retransmission is successful, re-queuing is not required. Therefore, equation \eqref{eq1} can be updated as:}  
\begin{equation}\label{eq13}   
		\textcolor{red}{T_q = \frac{2}{N_s} \sum_{m=1}^{N_s/2} \left\{ RRI + \frac{E[T_a]}{(1 - p_d^2)} + \frac{\max(t_{\text{GAP}}, RRI/2) p_d}{(1 - p_d)} \right\}.}
\end{equation}
\textcolor{red}{It can be quickly verified that when \(P_{coll}\) or \(P_d\) equals zero, the model degenerates to the case considering only collisions or only transmission failures, respectively.}

Assuming that the amount of data a vehicle needs to transmit is $\omega$, the data transmission delay can be calculated based on Shannon's formula as follows:
\begin{equation}\label{}   
    T_t = \frac{\omega }{{B{{\log }_2}(1 + \frac{{P \cdot G}}{{{N_0} \cdot B}})}},
\end{equation}
where $B$ is the bandwidth, $P$ is the transmit power, $G$ is the channel gain, and ${N_0}$ is the noise power spectral density. Therefore, the system's AoI can be expressed as:
\begin{equation}\label{}   
	{\rm A}  = {T_q} + {T_t}.
\end{equation}

\textcolor{red}{Based on the above analysis, we formulate the following optimization problem. The objective is to minimize the system AoI by jointly optimizing vehicle speed, vehicle density, and RRI:}
\begin{equation}\label{}
	\begin{aligned}
		\mathop{\min}\limits_{\rho,\,v,\,\mathrm{RRI}} &\ \mathrm{A}, \\
		\text{s.t.}\quad & \rho = \frac{Q}{v},\ Q \in \mathbb{R}, \\
		& \rho_{\min} \leq \rho \leq \rho_{\max}, \\
		& v_{\min} \leq v \leq v_{\max}, \\
		& \mathrm{RRI}_{\min} \leq \mathrm{RRI} \leq \mathrm{RRI}_{\max},
	\end{aligned}
\end{equation}
\textcolor{red}{where} \(\rho_{\min}\), \(\rho_{\max}\), \(v_{\min}\), \(v_{\max}\), \(\mathrm{RRI}_{\min}\), and \(\mathrm{RRI}_{\max}\) denote the feasible ranges of vehicle density, speed, and RRI, respectively. \textcolor{red}{Vehicle density and RRI primarily affect the collision probability, while vehicle speed influences the packet loss probability over the wireless channel. Transmission failures cause retransmissions and queue reordering, further increasing the AoI. Due to the coupling between vehicle speed and density under fixed traffic flow conditions, and since the AoI calculation involves multiple nonlinear function compositions, the optimization problem becomes highly complex. Consequently, balancing these environmental parameters to minimize the system AoI presents a significant challenge.}

\section{Optimization Strategy}
\textcolor{red}{
	In this section, we formulate the optimization problem as a Markov Decision Process (MDP) and solve it respectively using a pretrained LLM and the Deep Deterministic Policy Gradient (DDPG) algorithm.
}
\vspace{-0.2cm}

\subsection{DDPG Method}
\textcolor{red}{
	To apply the DDPG method, we first need to specify the state, action, and reward that define the MDP.
}

\subsubsection{State}
	In the environment considered in this paper, the key variables affecting the system's AoI are vehicle speed, vehicle density, and RRI. \textcolor{red}{Additionally, the collision probability \(P_{\text{coll}}\) and packet drop probability (\(p_d\)) act as intermediate variables that directly influence the AoI. Therefore, the system state at time \(t\) is defined as:
}
\begin{equation}\label{}
	\mathbf{s}_t = [v_t, \rho_t, \mathrm{RRI}_t, p_{d,t}, P_{\text{coll},t}],
\end{equation}
where \(v_t\) denotes the vehicle speed, and \(\rho_t\) denotes the vehicle density. The vehicle speed and density satisfy the constraints imposed by a fixed traffic flow scenario.

\subsubsection{Action}
The system's decision variables mainly include vehicle speed, vehicle density, and RRI. Since vehicle density can be derived from speed and a fixed traffic flow constant, the action space only includes vehicle speed and RRI, represented as:
\begin{equation}\label{}
	\mathbf{a}_t = [\mathrm{RRI}_{t+1}, v_{t+1}].
\end{equation}

\subsubsection{Reward}
The objective of the optimization problem is to minimize the system's AoI. To better incentivize the model to explore actions that yield improvements, we design a piecewise reward function with different slopes on different segments. The specific form is as follows:
\begin{equation}
	\small
	R = 
	\begin{cases}
		-\dfrac{\text{AoI}}{N_1} + N_2, \text{if } \text{AoI} > A_1, \\[6pt]
		\left(-\dfrac{A_1}{N_1} + N_2\right) + (A_1 - \text{AoI}), \text{if } A_2 < \text{AoI} \leq A_1, \\[6pt]
		\left(-\dfrac{A_1}{N_1} + N_2\right) + (A_2 - A_1) + 3 (A_2 - \text{AoI}), \text{if } \text{AoI} \leq A_2,
	\end{cases}
	\label{eq:reward}
\end{equation}
where \(N_1\) and \(N_2\) are scaling factors, and \(A_1\) and \(A_2\) are threshold points defining the piecewise segments.

\textcolor{red}{After defining the state, action, and reward, the problem can be formulated as a MDP, denoted as \(M = (\mathcal{S}, \mathcal{A}, P, R, \gamma)\), where \(P\) is the state transition probability and \(\gamma\) is the discount factor. The transition kernel \(P(s_{t+1} | s_t, a_t)\) is determined by the combination of \textcolor{red}{SPS access in the simulator} and the physical layer packet loss process. After executing an action, \textcolor{red}{the state is updated accordingly}, and the reward is obtained.}

\textcolor{red}{
	To obtain the optimal solution in continuous space, we employ the DDPG algorithm framework, which consists of the following components:
}
\textcolor{red}{
	\begin{itemize}
		\item A deterministic policy network (actor) \(\mu(s_t|\theta^\mu)\) that outputs actions \(a_t\), and a value network (critic) \(Q(s_t, a_t|\theta^Q)\);
		\item Target networks \(\mu'\) and \(Q'\) to improve training stability;
		\item An experience replay buffer to break sample correlations and enable uniform sampling;
		\item Ornstein–Uhlenbeck (OU) noise for exploration in continuous action spaces.
	\end{itemize}
}
\textcolor{red}{
	At each timestep, the agent executes an action:
	\begin{equation}
		a_t = \mu(s_t) + N_t,
	\end{equation}
	where \(N_t\) denotes the OU noise. \textcolor{red}{The transition tuple \((s_t, a_t, r_t, s_{t+1}, d_t)\), where \(d_t\) is the episode termination flag, is stored in the replay buffer.}
}
\textcolor{red}{
	During training, a minibatch of size \(N\) is randomly sampled from the buffer to update network parameters. The target value for the critic network is calculated as:
	\begin{equation}
		y_t = r_t + (1 - d_t) \gamma Q'(s_{t+1}, \mu'(s_{t+1})),
	\end{equation}
	The critic loss function is defined as the mean squared error:
	\begin{equation}
		L_Q = \frac{1}{N} \sum_t \big(Q(s_t, a_t) - y_t\big)^2.
	\end{equation}
}
\textcolor{red}{
	The policy network is updated by maximizing the state-action value function, using the deterministic policy gradient:
	\begin{equation}
		\nabla_{\theta^\mu} J \approx \frac{1}{N} \sum_t \nabla_a Q(s, a|\theta^Q) \big|_{a=\mu(s)} \nabla_{\theta^\mu} \mu(s|\theta^\mu).
	\end{equation}
}
\textcolor{red}{
	To ensure stable training, the target networks are softly updated as:
	\begin{equation}
		\theta' \leftarrow \tau \theta + (1 - \tau) \theta', \quad \tau = 0.005.
	\end{equation}
}
\begin{figure}[t]
	\centering 
	\includegraphics[width=\columnwidth]{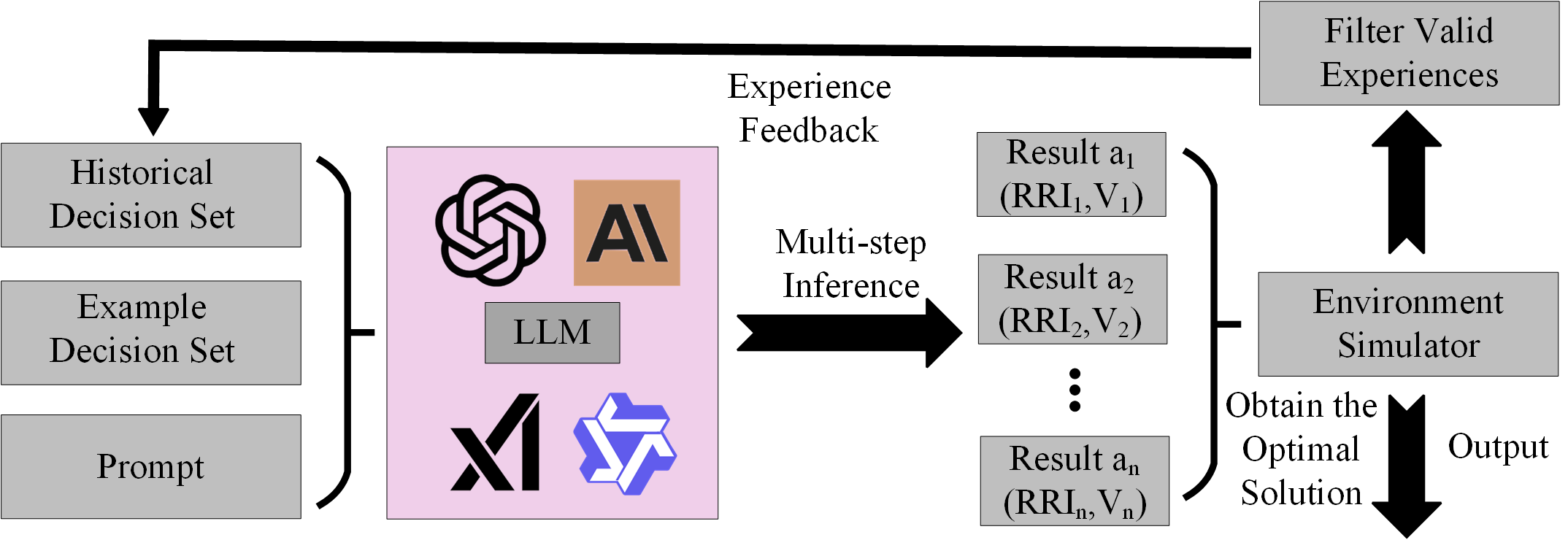}
	\caption{LLM Algorithm Framework.}
	\label{fig:5}       
	\vspace{-0.5cm}
\end{figure}
\vspace{-0.8cm}
\subsection{LLM-Based Approach}
LLMs are neural network-based language models with a large number of parameters, pre-trained on vast amounts of unlabeled text data~\cite{[15]}. As a result, they can understand complex language patterns~\cite{[16]} and demonstrate excellent performance across various natural language processing (NLP) tasks. Notably, LLMs possess in-context learning capabilities, enabling them to handle previously unseen tasks through textual guidance without any parameter fine-tuning\cite{888999}.

It is important to note that due to the massive parameter size, LLMs typically have relatively high response latency and are difficult to deploy. The problem addressed in this paper is an offline problem and does not require real-time local deployment, thus response latency is not a concern.

To clearly convey the task to the LLM, we divide the task description into five components: task background, task objective, main task, output format, and example sets. The task background explains the components of AoI and describes how vehicle density, RRI, and vehicle speed affect resource contention and transmission efficiency. The task objective specifies that the LLM should minimize system AoI by adjusting parameters. The main task defines the role of the LLM as an optimization algorithm, guiding it to iteratively infer parameters based on example decisions and historical decisions. The output format standardizes the presentation of results for subsequent processing. The example decision set provides representative parameter choices along with their corresponding AoI values to help the model determine initial solutions. The historical decision set supplies past parameters and their AoI results as references for the LLM inference. The detailed prompt can be obtained from the code (see the bottom right corner of the homepage).

\textcolor{red}{Fig.~\ref{fig:5} illustrates the overall framework of the LLM algorithm. At the beginning of each iteration, the decision data and prompt are fed into the LLM. To ensure the reliability of the output, the current input is frozen, and multi-step inference is performed using the LLM to generate multiple output results \(a_t\) (consistent with the actions in DDPG). The environment simulator calculates the AoI based on these results to obtain the current optimal solution. At the same time, duplicate experiences generated during the multi-step inference process are excluded, and the remaining valid experiences are added to the historical decision set as input data for the next iteration. When multiple inferences show no significant improvement, the model is considered to have reached convergence.}
\vspace{-0.2cm}
\section{Simulation}
\begin{figure*}[ht]
	\centering
	\subfloat[AoI vs Vehicle Speed for Different RRI.]{
		\includegraphics[width=0.31\textwidth, trim={0 0 0cm 0cm}, clip]{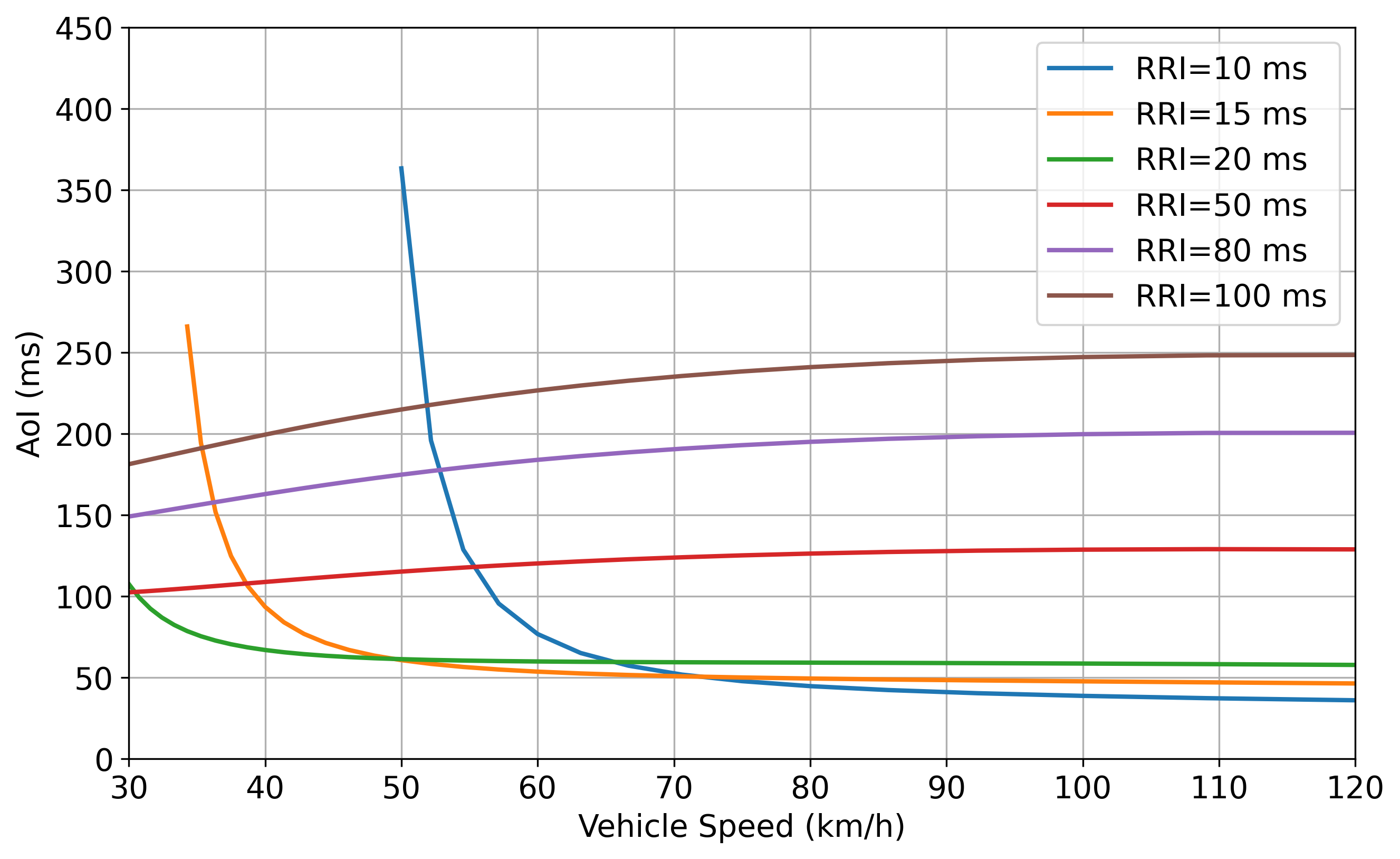}
		\label{fig:2}
	}
	\hfill
	\subfloat[AoI vs RRI for Different Vehicle Density.]{
		\includegraphics[width=0.31\textwidth, trim={0 0 0cm 0cm}, clip]{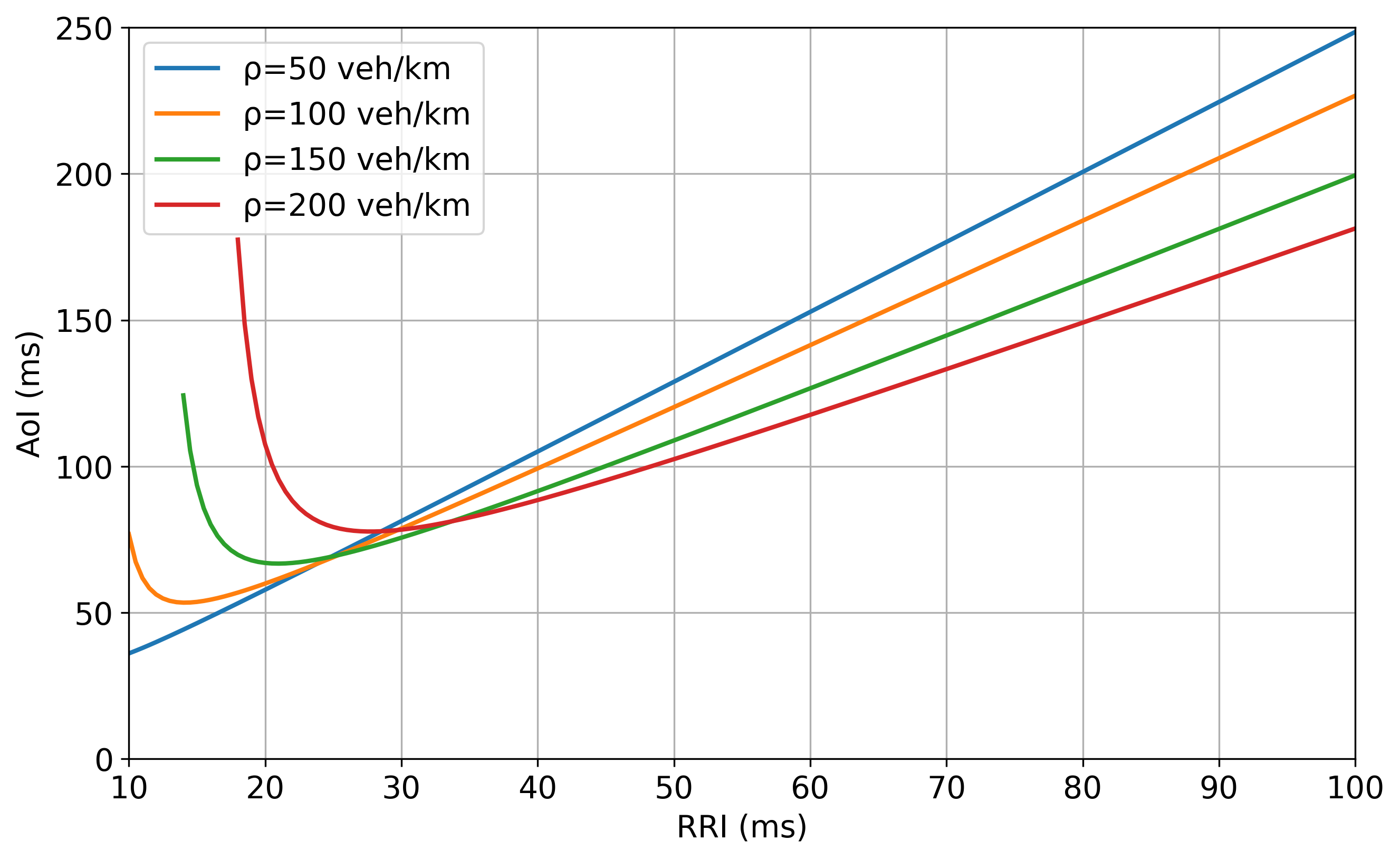}
		\label{fig:3}
	}
	\hfill
	\subfloat[AoI obtained by different methods.]{
		\includegraphics[width=0.31\textwidth, trim={0 0 1cm 1cm}]{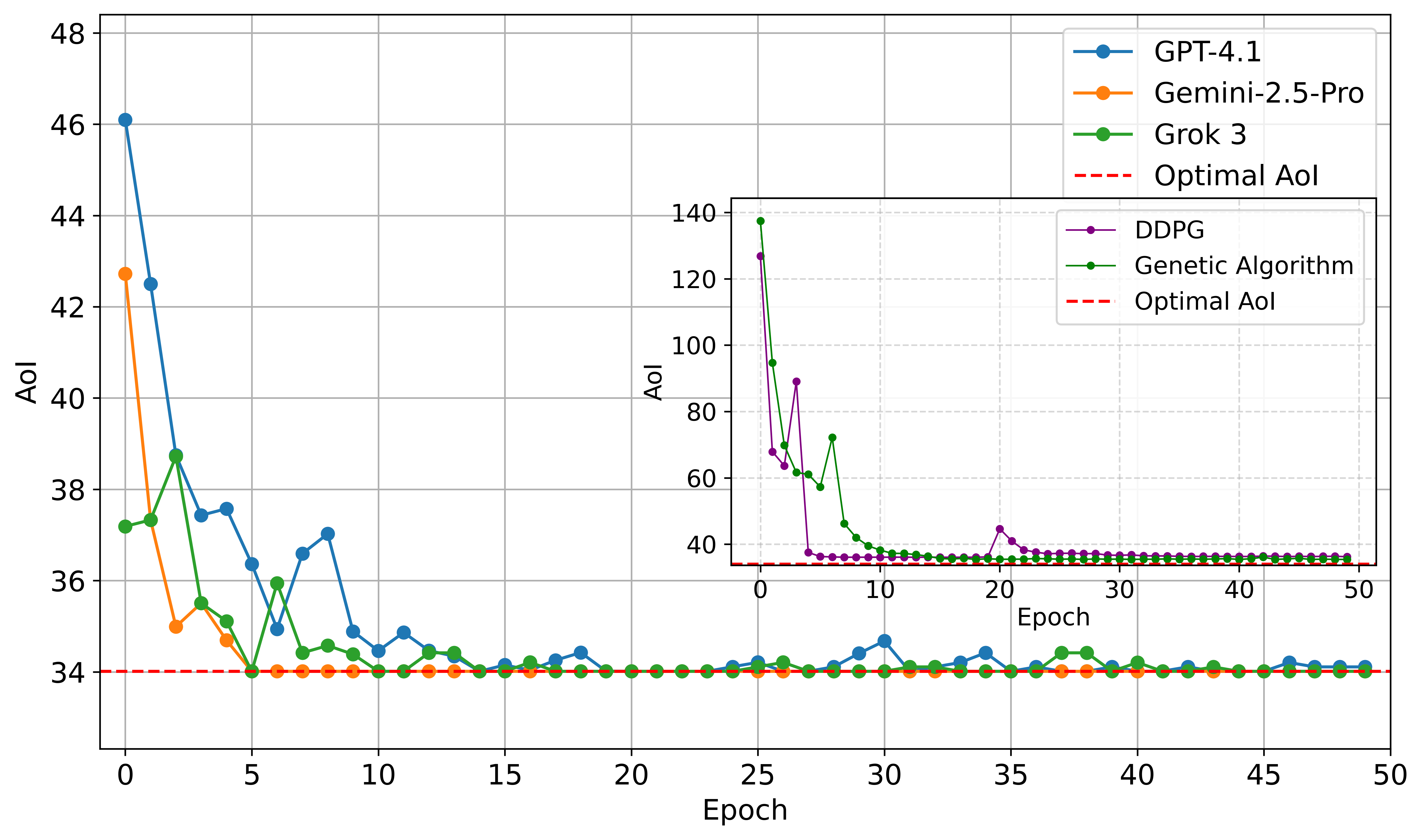}
		\label{fig:4}
	}
	\caption{Comparison of AoI under different conditions and methods.}
	\label{fig:combined}
	\vspace{-0.4cm}
\end{figure*}
To evaluate the effectiveness of the proposed optimization method, we conducted simulation experiments analyzing the impact of vehicle speed, RRI, and vehicle density on the AoI. The simulation is based on a highway scenario, assuming a traffic flow of 6000 vehicles per hour, \textcolor{red}{an RRI range of [10–100] ms, a vehicle density range of [50–200] vehicles per kilometer, and a speed range of [30–120] km/h.} The results are presented in three figures: Fig.~\ref{fig:2} and Fig.~\ref{fig:3} illustrate the functional relationships among RRI, vehicle speed, and AoI, while Fig.~\ref{fig:4} compares the performance of different methods on this problem.

\textcolor{red}{Fig.~\ref{fig:2} examines the impact of vehicle speed on AoI under varying RRI settings. Since the flow rate is kept constant, higher speeds correspond to lower densities. At low speeds (high density), AoI increases mainly due to a higher packet collision probability during SPS resource selection. In contrast, at high speeds (low density), although collisions are rare, severe Doppler-induced channel degradation raises the physical-layer failure rate, which leads to more blind retransmissions and consequently increases AoI. When RRI is large, the collision probability becomes negligible and transmission failures are primarily due to channel conditions, that is, speed. As a result, AoI gradually increases with speed. Conversely, when RRI is small, collisions dominate, especially under high density, making density the primary factor influencing AoI. Therefore, AoI peaks at low speeds (high density) and decreases as speed increases (density drops), eventually stabilizing. Additionally, the shorter retransmission interval with small RRI further improves AoI performance.}

Fig.~\ref{fig:3} shows the impact of RRI on AoI at different vehicle speeds, corresponding to the speed-density relationship in Fig.~\ref{fig:2}. The results indicate that at low vehicle speeds (high density), the collision probability is high at low RRI values, leading to larger AoI. As AoI increases, the relationship between RRI and AoI gradually becomes linear. This is because the collision probability stabilizes while the channel discard probability remains constant (due to fixed speed), making equation \eqref{eq13} effectively a linear function of RRI, \textcolor{red}{which is consistent with the findings in \cite{[11]}. Differing from \cite{[11]}, this study considers transmission failures and notes that at low vehicle densities with high vehicle speeds, the channel packet loss probability increases. Under these conditions, AoI increases with RRI at a faster rate than at low speeds, reflected by a steeper slope in the corresponding curves.}

Fig.~\ref{fig:4} compares the AoI variation trends over 50 training epochs among three large language models, the DDPG algorithm, and the genetic algorithm. The "Optimal" method represents the optimal value obtained through exhaustive search. Each data point is averaged over multiple runs. Due to the initial exploration of different parameter combinations by the LLMs, the AoI fluctuates significantly at the beginning. The prompts contain representative examples that provide good prior knowledge; therefore, the LLM methods have relatively good initial solutions and can converge quickly with only a few iterations. \textcolor{red}{The performance differences among various large-scale models are related to their model size and pre-training methods, which result in differing exploration strategy preferences.} In contrast, the DDPG and genetic algorithms require more steps of exploration in each epoch and need continuous training and adjustment of network parameters. The LLM, however, only needs to collect examples, making the algorithm implementation simpler.

\vspace{-0.2cm}
\section{Conclusion}
This letter proposes an AoI analysis model for IoV scenarios by coupling SPS queuing delay with a vehicle speed-sensitive channel failure mechanism. Experimental results validated that the joint optimization of vehicle speed, density, and RRI can significantly reduce AoI:  
\begin{enumerate}
	\item The LLM-based approach achieves rapid convergence driven by exemplars without requiring online training;
	\item The DDPG method demonstrates better stability in continuous space exploration but depends heavily on reward function design;
	\item LLMs equipped with certain prior knowledge provide better initial solutions, resulting in faster convergence compared to DDPG.
\end{enumerate}

Future research directions include designing lightweight fine-tuning schemes for LLMs to reduce dependency on exemplars, integrating the optimization engine into C-V2X roadside units to realize dynamic RRI configuration, and considering multi-base station scenarios to address AoI spikes caused by signal handovers.








\end{document}